\documentclass[lettersize,journal]{IEEEtran}
\usepackage{amsmath,amsfonts}
\usepackage{algorithmic}
\usepackage{algorithm}
\usepackage{array}
\usepackage[caption=false,font=normalsize,labelfont=sf,textfont=sf]{subfig}
\usepackage{textcomp}
\usepackage{stfloats}
\usepackage{url}
\usepackage{verbatim}
\usepackage{graphicx}
\usepackage{cite}
\usepackage{stackengine}
\usepackage{amssymb}
\usepackage{color}
\usepackage{bm}

\def\BibTeX{{\rm B\kern-.05em{\sc i\kern-.025em b}\kern-.08em
		T\kern-.1667em\lower.7ex\hbox{E}\kern-.125emX}}

\begin{document}

\title{
 Selective Experience Sharing in Reinforcement Learning Enhances Interference Management 
}

\author{Madan Dahal,~\IEEEmembership{Graduate~Student~Member,~IEEE}, and~Mojtaba~Vaezi,~\IEEEmembership{Senior~Member,~IEEE} 

 \thanks{This work was supported by the U.S. National Science Foundation under Grant CNS-2239524. 
 The authors are with the Department of Electrical and Computer Engineering, Villanova University, Villanova, PA, USA (E-mail:\{mdahal, mvaezi\}@villanova.edu).}%
}

	\maketitle
	

 \begin{abstract}

We propose a novel multi-agent reinforcement learning (RL) approach for inter-cell interference mitigation, in which agents selectively share their experiences with other agents.  Each base station is equipped with an agent, which receives signal-to-interference-plus-noise ratio from its own associated users. This information is used to evaluate and selectively share  experiences with neighboring agents. The idea is that even a few pertinent  experiences from other agents can lead to  effective  learning.
This approach enables fully decentralized training and execution, minimizes information sharing between agents and significantly reduces communication overhead, which is typically the burden of interference management. The proposed method outperforms state-of-the-art multi-agent RL techniques where training is done in a decentralized manner. Furthermore, with a 75\% reduction in experience sharing, the proposed algorithm achieves 98\% of the spectral efficiency obtained by algorithms sharing all experiences.

	\end{abstract}


	\section{introduction} \label{sec:intro}

Interference poses a significant challenge to achieving high throughputs and spectral efficiency in multi-cell cellular networks. Interference management has been extensively studied in the literature \cite{jafar2011interference,Sun2013InterferenceMT,el2013practical}, prompting research on various techniques, including interference alignment \cite{jafar2011interference} and coordinated multi-point  \cite{Sun2013InterferenceMT}. While these methods hold promise, their widespread adoption in wireless standards faces obstacles due to their high reliance on sharing data, control information, and channel state information between base stations (BSs). Such a need makes them ineffective in practical applications \cite{el2013practical}. Inter-cell interference coordination (ICIC) mitigates inter-cell interference by enabling coordination among BSs to improve \textit{signal-to-interference-plus-noise ratio} (SINR) via muting nearby interference \cite{zhang2014stochastic}. This reduces spectrum efficiency and capacity. In \cite{mei2019cooperative}, cooperative beamforming is used for distributed interference management in unmanned aerial vehicles.

\begin{figure*}[htbp] 
	\centering
	{
	\includegraphics[scale=.23, trim=0 0 0 0, 
		clip]{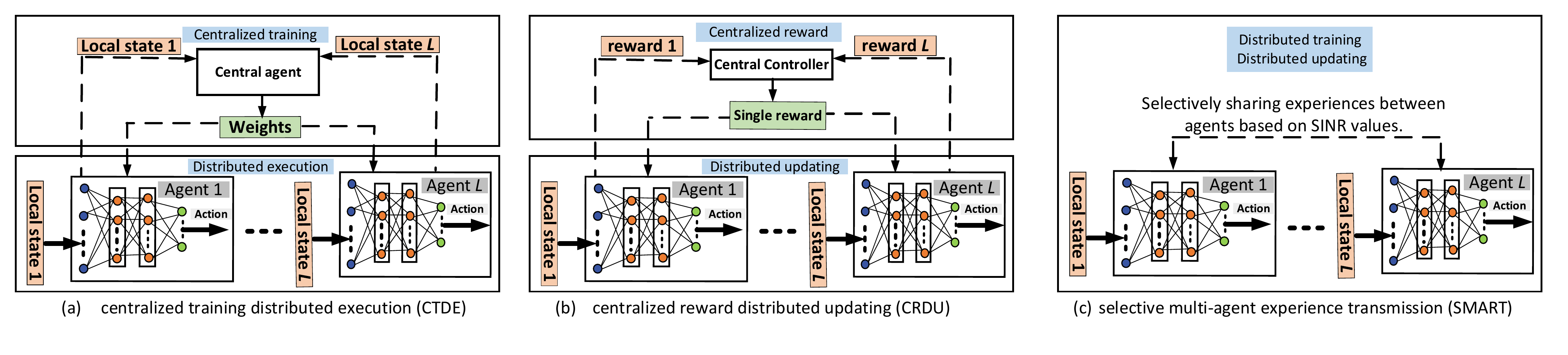}\label{fig_localCSI_a1a}}\vspace{-2.1mm}
 \caption{Comparison of three different multi-agent RL frameworks. (a) CTDE: agents receives updated weights of neural network from a central node, (b) CRDU: agents receive a common reward from central controller to update their network individually, and (c) SMART: agents receive selective experiences during training, eliminating reliance on a central agent or controller.}	\label{fig:MARL} 
\end{figure*}

Multi-agent reinforcement learning (RL) \cite{busoniu2008comprehensive} offers significant potential for inter-cell interference managingent with minimal communication overhead. In a multi-cell network, each cell is equipped with an agent capable of interacting with the environment by taking actions to maximize rewards, such as spectral efficiency or other desired metrics. Each agent operates independently, with access only to its local environment, allowing individual decision-making. While execution in multi-agent RL is distributed, training can be done in various forms, including centralized or decentralized manners, as shown in Fig.~\ref{fig:MARL} and discussed in the following.

Multi-agent RL has been applied to interference problem in various settings \cite{zhang2021joint,lu2021dynamic,liu2023double,zhang2022multi,dahal2023multi}. In \cite{zhang2021joint,lu2021dynamic,liu2023double}, a \textit{centralized training distributed execution (CTDE)} framework is used to maximize the sum-rate of the network.  However, the process of sharing local experiences with a central location for training, as well as transmitting neural network weights to each agent, results in significant communication overhead. In \cite{zhang2022multi,dahal2023multi}, a \textit{centralized
reward distributed updating (CRDU)} framework is used  to maximize the system sum-rate.  In this approach, the network update/training is performed locally. However, a central controller dictates rewards/penalties uniformly across all agents. This centralized reward can be limiting, as poor performance from one agent affects all. An alternative method involves agents with fully decentralized decision-making (distributed training, reward, and execution) but sharing their entire experiences, including state, action, and reward, with all other agents \cite{christianos2020shared}.  
While effective in interference mitigation, this strategy also incurs high communication overhead. 

We present a novel fully-distributed multi-agent RL approach for inter-cell interference management,  in which agents share a selected number of their experiences. The idea is that if one agent finds critical experiences in the environment, sharing them with other agents could help learning process. However, it is crucial to only share important experiences, as sharing all experiences increases complexity and communication overhead.  Each BS receives SINR from
its own associated users within its respective cell. This information is used to calculate the inter-cell interference power value and compare it with a threshold. If the calculated power is higher than the threshold, the corresponding \textit{experience} (state, action, and reward) is selected to be shared with other agents.

We name the proposed multi-agent RL  approach  \textit{selective multi-agent experience transmission (SMART).} In this framework, agents use a deep Q-network (DQN)-based algorithm individually for learning, and share their experiences based on the SINR values and interference level of their associated users. The advantages of SMART learning framework, compared to CTDE and CRDU approaches, shown in Fig.~\ref{fig:MARL}, are:

\begin{enumerate}
\item   This learning reduces communication overhead as it requires
only selective experiences to share among agents.  
    \item Learning rate will be faster as highly relevant experiences is shared with  higher performance.

\end{enumerate}

The goal of SMART is to maximize
the spectral efficiency of the network, manifested by network
sum-capacity. Simulation results confirm that  SMART performs significantly better than multi-agent RL without sharing experiences and is almost as effective as sharing all experiences between BSs.

	\section{System Model} \label{sec:model}
Consider a downlink cellular
    network with $L$ cells and $U$ user equipments (UEs) in each cell. Each BS {simultaneously} serves multiple single-antenna UEs and each UE can only be served by one BS at a time. Each BS is equipped with $M$ antennas in a \textit{uniform linear array}.  Due to hardware limitations on large-scale multiple-antenna systems, the BSs often use pre-defined beamforming codebooks \cite{zhang2017codebook} that scan all potential directions for data transmission.
    For simplicity, each beamforming vector's weights are implemented using constant-modulus $r$-bit quantized phase shifters. Beamforming vectors are selected from the codebook {whose} each element is given by $\mathbf{w}   = {\dfrac{1}{\sqrt{M}}}{\left[ e^{j\theta_1},\dots,  e^{j\theta_M} \right ]^T} $. 
	The phase shift $\theta_m, m = \{1,2, \dots, {M}\}$, is selected from a finite set $\mathbf\Phi$ with $2^r$ possible discrete values uniformly drawn from ${\left[0,\pi \right]}$.

    The transmitted signal from the $j$th BS at time step $t$ is given by $
 {\mathbf{x}_{j}} = \sum_{u=1}^U\mathbf{w}_{j,u}s_{j,u}
$. Here $\mathbf{w}_{j,u} \in \mathbb{C}^{M \times 1}$ is the beamforming vector for $u$th UE at $j$th BS and $s_{j,u}$ denotes the transmitted
data intended for $u$th UE with {$\mathbb{E}{[|\mathbf{s}_{j,u}|^2]} = P_{j,u}$, where $P_{j,u}$ being the power of  $j$th BS allotted to $u$th UE}. Also {$\mathbb{E}{[|\mathbf{x}_{j}|^2]} = P_{j}$}, where $P_{j}$ represents the transmit power from $j$th BS. Then the received signal at $u$th UE at {$\ell$}th cell is 
\vspace{-0.1cm}
    \begin{align}\label{eq:received signal}
y_{\ell,u}   = \mathbf{h}^H_{\ell,\ell,u}\mathbf{w}_{\ell,u} s_{\ell,u}+\sum_{k\neq u} \mathbf{h}^H_{{\ell},\ell,u}\mathbf{w}_{\ell,k}s_{\ell,k}  \notag \\ + \sum_{j\neq \ell}\sum_{u=1}^U \mathbf{h}^H_{{\ell},j,u}\mathbf{w}_{j,u}s_{j,u} + n_{\ell,u},
\end{align}
where $\mathbf{h}_{\ell,j,u} \in \mathbb{C}^{M \times 1}$,  $\ell, j \in \{1,\dots,L\}$, is the channel vector adopting the geometric channel
model \cite{el2014spatially} from $j$th BS to the $u$th UE in $\ell$th cell as described in  \cite[equation (3)]{mismar2019deep}, and $n_{\ell,u} \in \mathcal{CN}(0,\,\sigma^{2})$ is the noise at the $u$th UE with zero mean and variance of $\sigma^{2}$. The SINR of $u$th UE at $\ell$th cell  is
\begin{align}\label{eq:SINR} 
 \gamma_{\ell,u}  
= \frac{S_{\ell, u}}{\sigma^{2}+I_{\ell, u}^{{\rm Intra}}+ I_{\ell, j, u}^{{\rm Inter}}},
\end{align}	
where   $S_{\ell, u} \triangleq P_{\ell,u}|\mathbf{h}_{\ell, \ell, u}^{\mathrm{H}} \mathbf{w}_{\ell, u}|^{2}$ is the signal power, $I_{\ell, u}^{{ \rm Intra}} \triangleq \sum_{k\neq u}P_{\ell,k}|\mathbf{h}_{\ell,\ell,u }^H\mathbf{w}_{\ell,k}|^2$ is the intra-cell interference power experienced by $u$th UE served by $\ell$th cell, and $I_{\ell, j, u}^{{\rm Inter}} \triangleq \sum_{j\neq \ell}\sum_{u=1}^U P_{j,u}|\mathbf{h}_{\ell,j,u}^H\mathbf{w}_{j,u}|^2$ is the inter-cell interference experienced by $u$th UE at $\ell$th BS. The total interference power at $u$th UE served by $\ell$th BS is $I_{\ell, j, u}^{\rm Total} = I_{\ell, u}^{{\rm Intra}}+I_{\ell, j, u}^{{\rm Inter}}$.

The sum achievable rate, or simply sum-rate, is a common measure of spectral efficiency in cellular
networks. Considering this, in this paper our goal is to maximize the network sum-rate which is defined as ${\sum_{\ell = u }^U\log_2(1+\gamma_{\ell,u})}$, and is equivalent to   $\log_2 \prod(1+\gamma_{\ell,u})$. Since the logarithm is a monotonic function, to find the arguments that maximize the sum-rate we can solve 
\vspace{-0.1cm}
\begin{align}\label{eq:Optimization_problem}
{\max_{{\substack{P_{\ell,u}, \mathbf{w}_{\ell,u}}}}\quad} &
\prod_{u =  1}^U{(1+\gamma_{\ell,u})}\\
\textrm{subject to} \quad & P_{\ell,u}\in \mathcal{\bm P}, \mathbf{w}_{\ell,u}\in \mathcal{\bm W},\quad\forall\ell, \forall u,\\
& \sum_{u} P_{\ell,u} \leq { P_\ell^{\rm max}},{\gamma}_{\ell,u} \geq  {\gamma_{\rm min}},\quad\forall \ell, \forall u,
\end{align}
in which $\bf \mathcal {W}$  is   \textit{beamforming codebook} from which $\mathbf{w}_{\ell,u}$ is selected,
  $\bf \mathcal {P}$ is the possible transmit powers, $P^{\rm max}_{\ell}$ is the maximum power for $\ell$th BS, and $\gamma_{\rm min}$ denotes the  the minimum SINR for any UE to guarantee their quality of service requirements. The problem \eqref{eq:Optimization_problem} is challenging and non-convex, and traditional methods have limitations, including computational complexity and adaptability to evolving environments.

  \begin{figure*}[htbp] 
	\centering
	{
		\includegraphics[scale=.15, trim=0 0 0 0, 
		clip]{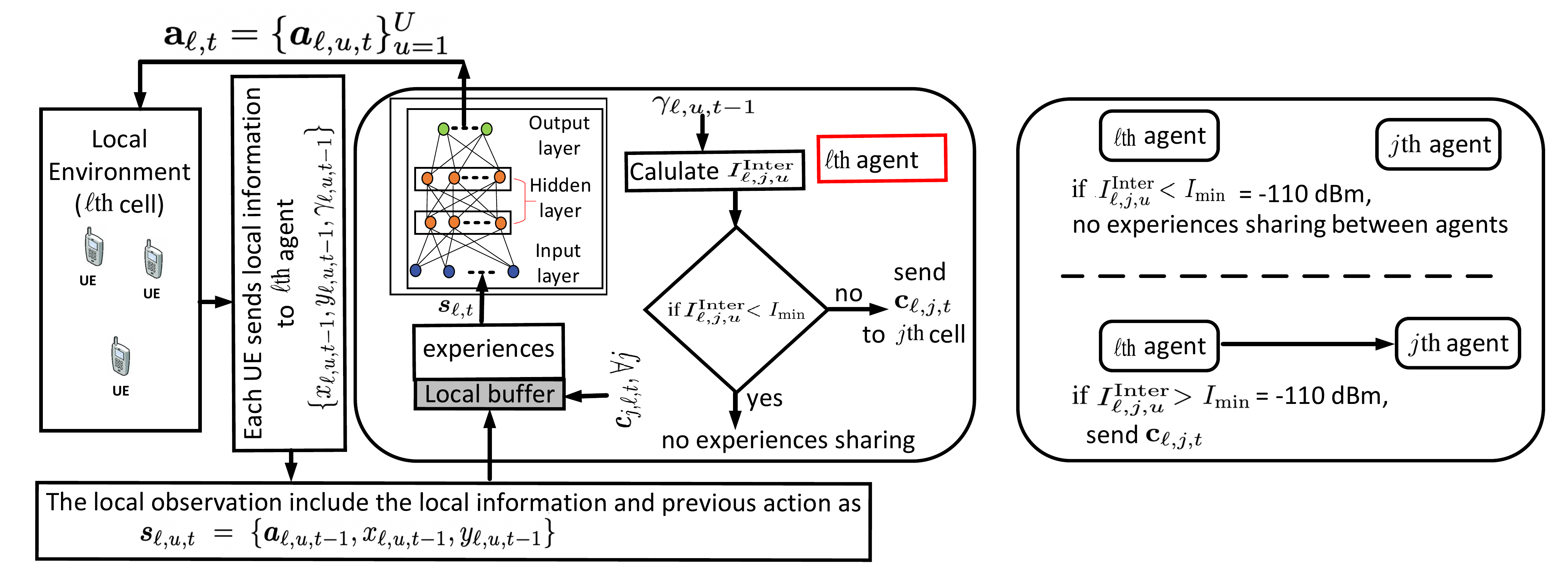}\label{fig_localCSI_a1a}}\hspace{-2.1mm}
	\caption{ The
schematic shows the agent architecture and the way it interacts
with the environment. (left) An illustration of the proposed SMART system, and (right) details of the communication between the agents.}
	\label{fig:cmpSNRR} 
\end{figure*}  

Our approach to solve this problem is described below.  At the beginning of time step $t$, $\ell$th BS uses the transmit power and beamforming vectors of the previous time step to determine the serving power at $u$th UE served by $\ell$th BS as
\begin{align}\label{eq:serving_power}
S_{\ell, u,t}=P_{\ell,u,t-1}|\mathbf{h}_{\ell, \ell, u,t-1}^{\mathrm{H}} \mathbf{w}_{\ell, u,t-1}|^{2}.
\end{align}
Also, intra-cell interference power at $u$th UE served by $\ell$th BS at time step $t$ is evaluated  as
\vspace{-0.2cm}
\begin{align}\label{eq:intra_power}
I_{\ell, u,t}^{{\rm Intra}}=\sum_{k\neq u}P_{\ell,k,t-1}|\mathbf{h}_{\ell,\ell,u,t-1 }^H\mathbf{w}_{\ell,k,t-1}|^2.
\end{align}	

We proposed using SINR measured by UEs for this purpose. We should highlight that starting with 5G New Radio (NR)  \cite{vaezi2024deep,3gpp.38.215}, SINR (i.e., $\gamma_{\ell,u}$) can be measured directly by UEs and reported to their serving BS. Thus, based on reported  $\gamma_{\ell,u}$, using  \eqref{eq:SINR},  we measure  $I_{\ell, j, u}^{\rm Total}$ + $\sigma^2$  at time step $t$ as 
\vspace{-0.2cm}
\begin{align}\label{eq:total_interfernce_power}
I_{\ell, j, u,t}^{\rm Total}+\sigma^2= S_{\ell, u,t}/ \gamma_{\ell,u}.
\end{align}
By subtracting noise power from this value gives $I_{\ell, j, u}^{\rm Total}$. Then the  inter-cell interference power, $I_{\ell, j, u}^{{\rm Inter}}$ from the $j$th BSs at $u$th UE served by $\ell$th BS at time step $t$ is obtained as
\begin{align}\label{eq:inter_power}
I_{\ell, j, u,t}^{{\rm Inter}}= I_{\ell, j, u,t}^{\rm Total}-I_{\ell, u,t}^{{\rm Intra}}.
\end{align}

Based on $I_{\ell, j, u,t}^{{\rm Inter}}$, selective  experiences are shared between BSs. Specifically, only if $I_{\ell, j, u,t}^{{\rm Inter}}$ exceeds $I_{\min} = -110$~dBm, the minimum interference threshold,  experiences are shared between BSs. We selected this value heuristically, knowing that cellphone's SINR sensitivity is approximately -110 dBm, which corresponds to the noise level. This means that if interference is lower than the noise level, it can be disregarded. This would greatly reduce communication overhead as selective experiences are shared between
BSs.  In general, the threshold value could be adjusted based on noise floor, interference strengths, and UE sensitivity. Throughout the above process, we assume that each BS will only have its associated UEs CSI and SINR, which are measured and reported \textit{locally}. 
In the following, we detail the design of our algorithm that selectively shares experiences between BSs.

\section{SMART: Selective Multi-Agent Experience Transmission}\label{sec:DLnet}

In our approach, agents selectively share experiences with each other. The idea is that not all experiences are needed to be shared to agents to discovers significant insights of the environment, sharing only critical experiences with other agents can accelerate their learning process yet to have comparative performance.   
The steps at each agent are as follows:
\begin{itemize}
    \item  collect local experiences and store in a \textit{local} replay buffer.
    \item   share  experiences  with other agents if inter-cell interference power  $I_{\ell, j, u}^{{\rm Inter}}$ satisfies certain conditions. 
    \item insert received experiences (if any) in  replay buffer.
    \item   sample a minibatch of experiences from their own replay buffer and perform gradient descent (GD).
\end{itemize}

We note that the agents only interact during the experience sharing which hugely reduces the communication overhead.
The local state observed by the $\ell$th agent is $\mathbf{s}_{\ell,t} = \{\bm s_{\ell,u,t}\}_{u=1}^U$, where $\bm{s}_{\ell,u,t} = \{\bm{a}_{\ell,u,t-1}, x_{\ell,u,t-1}, y_{\ell,u,t-1}\}$. Here $x_{\ell,u,t-1}$ and $y_{\ell,u,t-1}$ are the coordinates of the $u$th UE in the $\ell$th cell, $\bm{a}_{\ell,u,t-1} = \{P_{\ell,u,t-1},\mathbf{w}_{\ell,u,t-1}\}$ is the previous action. By keeping track of the UE's coordinates using reliable localization methods like satellite navigation and 3-dimensional ranging \cite{yang2021driving}, the network can make better informed decisions, which results in improved performance \cite{mismar2019deep}. The interference coordination and power control for $u$th UE at $\ell$th BS is
\vspace{-0.2cm}
\begin{equation}
\begin{aligned} \label{eq:power command}
  P_{\ell,u,t} := 
      {{   {P_{\ell,u,t-1}+PC_{\ell,u,t}} }}, 
\end{aligned}
\end{equation}
in which
	$PC_{\ell,u,t}$ is the power control command for the $u$th UE at $\ell$th BS which is $+ 1$dB or $-1$dB depending on  the action related to that command. If $\sum_{u=1}^U P_{\ell,u,t} > P^{\rm max}_\ell$, then $PC_{\ell,u,t}$ will be pushed to $-1$dB to obey the total power limit. The string of bits with the help of bitwise-AND and shifting enables joint actions concurrently. Specifically, for any $u$th UE in $\ell$th BS we have
\vspace{-0.3cm}
 \begin{align}\label{eq:action}
	{\bm{a}_{\ell,u,t}}   = { \{\underbrace{a_{\ell,u,t}^1}_\text{power control}, \underbrace {a_{\ell,u,t}^2}_\text{beamforming}\}}.
	\end{align}
 
 where action $a_{\ell,u,t}^1$ adjusts the transmit power of the $u$th UE in the $\ell$th BS: $a_{\ell,u,t}^1 = 0$ decreases power by 1 dB, while $a_{\ell,u,t}^1 = 1$ increases it by 1 dB. Similarly, $a_{\ell,u,t}^2$ modifies the beamforming codebook index: $a_{\ell,u,t}^2 = 0$ steps it down, and $a_{\ell,u,t}^2 = 1$ steps it up. Action $\bm a_{\ell,t}$ taken by $\ell$th agent is a binary vector of
length $2U$ and has the following form, ${\mathbf a_{\ell,t}}   = { \{{\bm{a}_{\ell,u,t}} \}_{u=1}^U}$.
The agent’s final objective
is to maximize the total cumulative reward which is defined as
\begin{align} \label{eq:rew_1}
  {r_{\ell,t}} =
    \begin{cases}
      \prod \limits_{u =  1}^U{(1+  \gamma_{\ell,u})}, & \; \text{if}\; 
  \gamma_{\ell, u,t} >  \gamma_{\rm min}\;\text{and}\;I_{\ell, j, u}^{{\rm Inter}} < I_{\rm min},\\
      -\Re, & \text{otherwise,}
    \end{cases}      
\end{align}
where $\Re$ is a positive constant that acts as the punishment, making the agent explore the environment more. The $\ell$th agent stores experiences, $ E_{\ell,t}$  in local buffer as 
\vspace{-0.1cm}
\begin{align}\label{eq:action_1}
 \hspace{-0.12cm} E_{\ell,t} \hspace{-0.12cm} = \hspace{-0.12cm} 
\begin{Bmatrix}
\bm e_{\ell,u,t}, \\
\vdots \\
\bm e_{\ell,U,t} 
\end{Bmatrix}\hspace{-0.12cm}=\begin{Bmatrix}
\bm s_{\ell,u,t}, & \bm{a}_{\ell,u,t}, & r_{\ell,t}, &\bm s_{\ell,u,t+1} \\
\vdots & \vdots &\vdots & \vdots \\
\bm s_{\ell,U,t}, & \bm{a}_{\ell,U,t}, & r_{\ell,t}, & \bm s_{\ell,U,t+1} 
\end{Bmatrix}
\end{align}

The $\ell$th agent selectively share the experiences, $\bm c_{j,\ell,t} $ to the $j$th agents based on the  inter-cell interference power, $I_{\ell, j, u}^{{\rm Inter}}$ as calculated in \eqref{eq:serving_power} to \eqref{eq:inter_power}. Mathematically,
\begin{equation}
\begin{aligned} \label{eq:comm}
  {\mathbf c_{j,\ell,t}} =
    \begin{cases}
      \mathbf e_{\ell,u,t}, & I_{\ell, j, u,t}^{{\rm Inter}} > I_{\min}, \forall u,\\
      \text{no  experience shared}, & I_{\ell, j, u,t}^{{\rm Inter}} \leq I_{\min},\forall u
    \end{cases}      
\end{aligned}
\end{equation}

The shared experiences are inserted in the agent's local replay buffer. At each training step, a $B$ mini-batch of experiences are sampled from a replay buffer. Let $b = \langle \mathbf{s}_{b}, \mathbf{a}_{b}, r_{b}, \mathbf{s}'_{b} \rangle$ denote an experience in the mini-batch $B$ from $\ell$th agent local buffer. Then the loss function of the DQN network of $\ell$th agent with the initial weight $\bm\theta_t$   is given as
\vspace{-0.2cm}
 \begin{align}\label{eq:Loss}
	{ L({\bm\theta_t})} = \dfrac{1}{B}\sum_{b = 1 }^B{\left[(y_b-Q_\ell(\mathbf{s}_{b},\mathbf{a}_{b};{\bm \theta}_t))^2\right]},
	\end{align}	
where $y_b = r_{b}+\alpha \max_{\mathbf{a}'_{b}}Q_\ell (\mathbf{s}'_{b},\mathbf{a}'_{b};{\bm \theta}_{t-1})$, $Q_\ell(\mathbf{s}_{b}, {\mathbf{a}_{b}};\bm{\theta}_{t})$ is state-action value function that describes the
expected reward after taking one specific action following
the policy $\pi$ and $\alpha$ is a discount factor {whose} range is $[0,1]$. The gradient of loss function  with respect to $\bm\theta_t$ is taken. In every iteration, the weight $\bm\theta_t$ is updated based on the gradient of the loss function. The update rule for $\bm\theta_t$ is 
$\bm\theta_{t+1} = \bm\theta_t - \eta \nabla_{\bm\theta_t} L(\bm\theta_t)$, where $\eta$ is the learning rate.
The ultimate goal of updating the weight $\bm\theta_t$
in every iteration is to minimize the loss function \eqref{eq:Loss} of the DQN network. 
Let $I$, $h_1$, $h_2$, and $O$ represent the sizes of the input, hidden, and output layers, respectively. The action dimension is $2U$. The total number of parameters is $I + h_1 + h_2 + O$, and the complexity is $\mathcal{O} (2U(I + h_1 + h_2 + O))$.

By using shared experiences, each agent develops a more comprehensive state-action value function. This enables the agent to better predict how its actions affect the users, both in its own cell and neighboring cells, leading to improved action selection and reduced interference for users in adjacent cells.

\begin{figure*}[!t]
\centering
\subfloat[]{\includegraphics[width=2.0in]{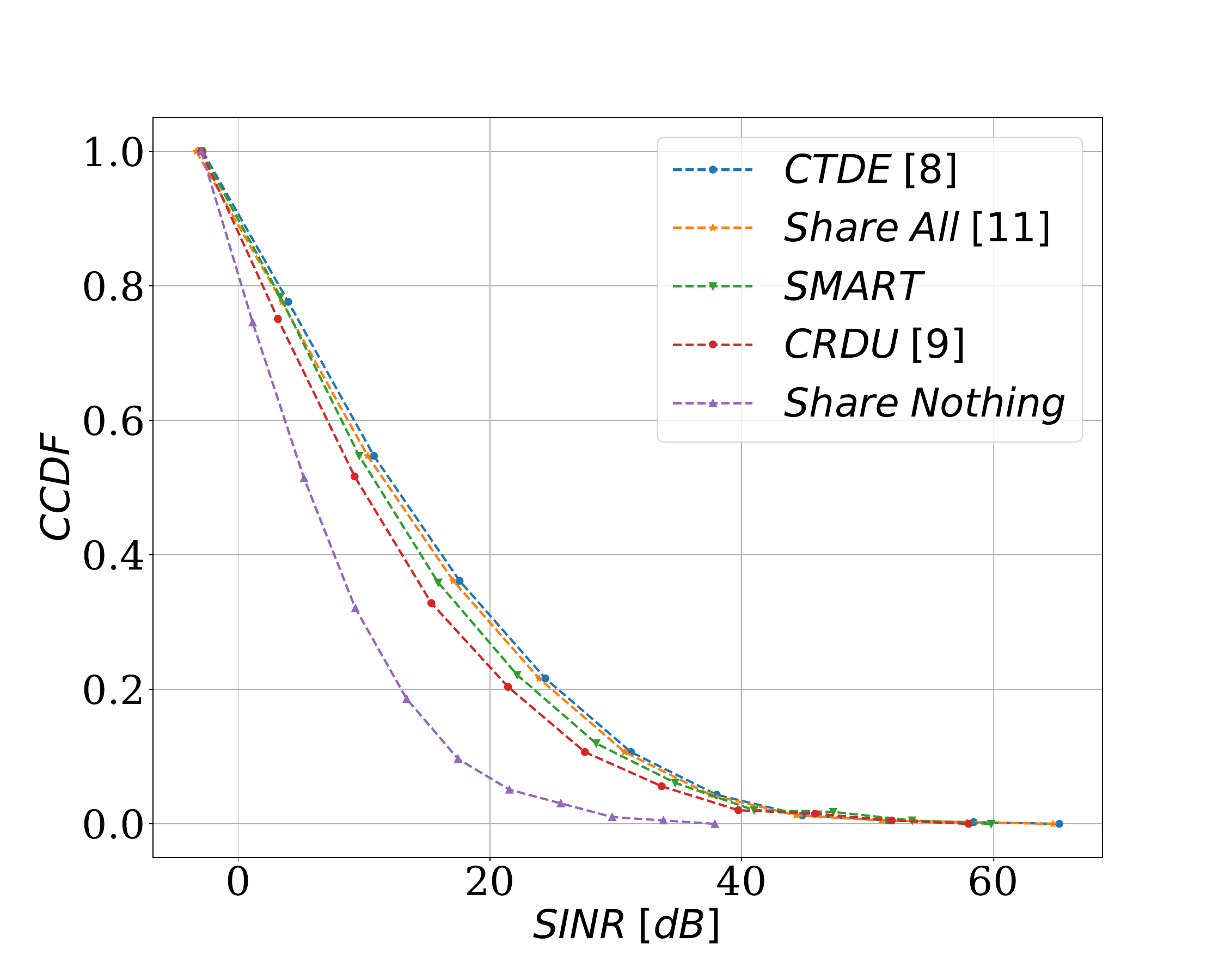}%
\label{a}}
\hfil
\subfloat[]{\includegraphics[width=2.57in]{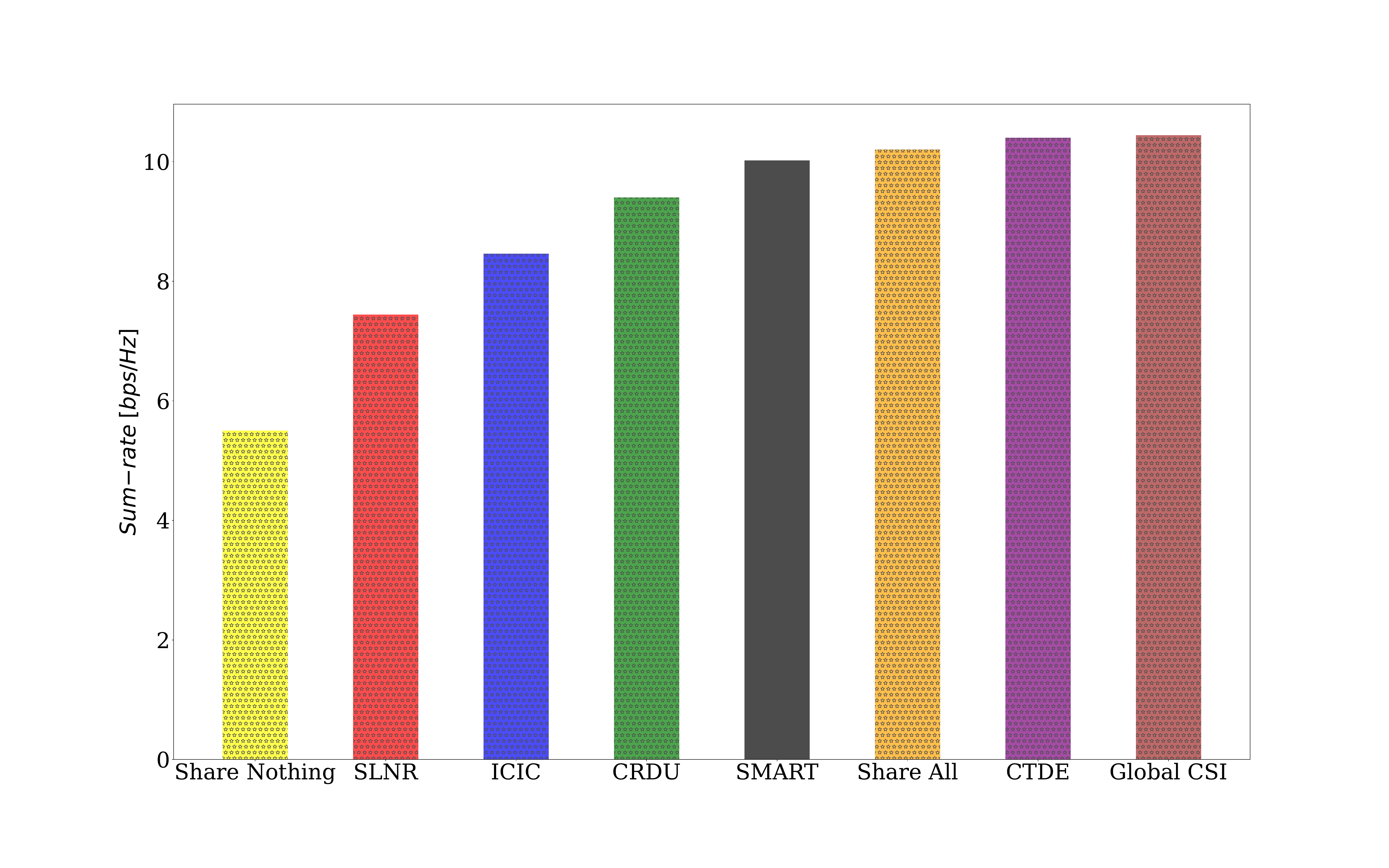}%
\label{b}}
\subfloat[]{\includegraphics[width=2.52in]{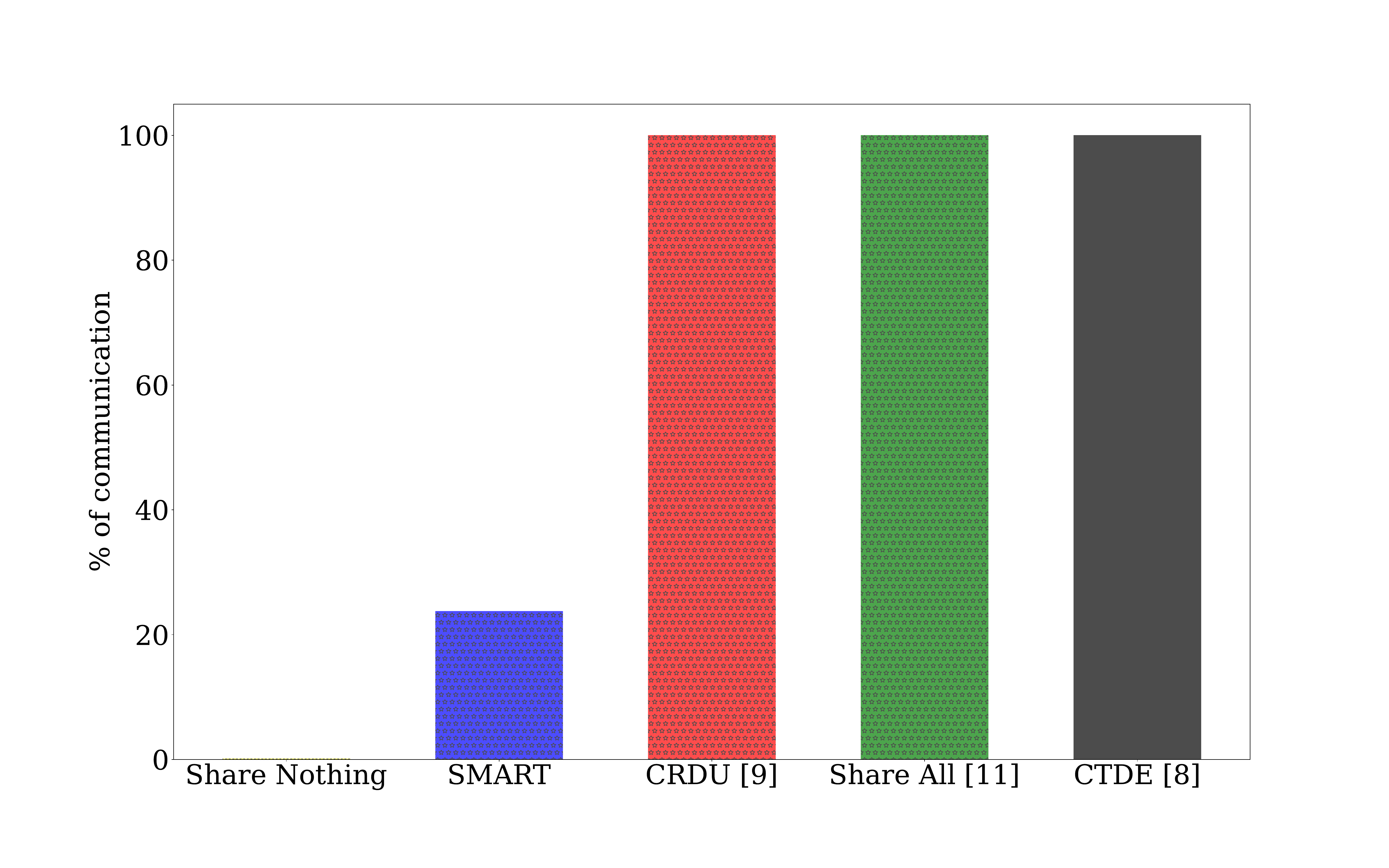}%

\label{c}}
\hfil
\caption{Performance of the SMART algorithm versus others ($L=2$). (a) CCDF of the  SINR values, (b) network sum-rate, (c) communication overhead.}
\label{fig:cmpSNRR1}
\end{figure*}

\begin{figure}[tbp]
		\centering
\includegraphics[width=0.32\textwidth]{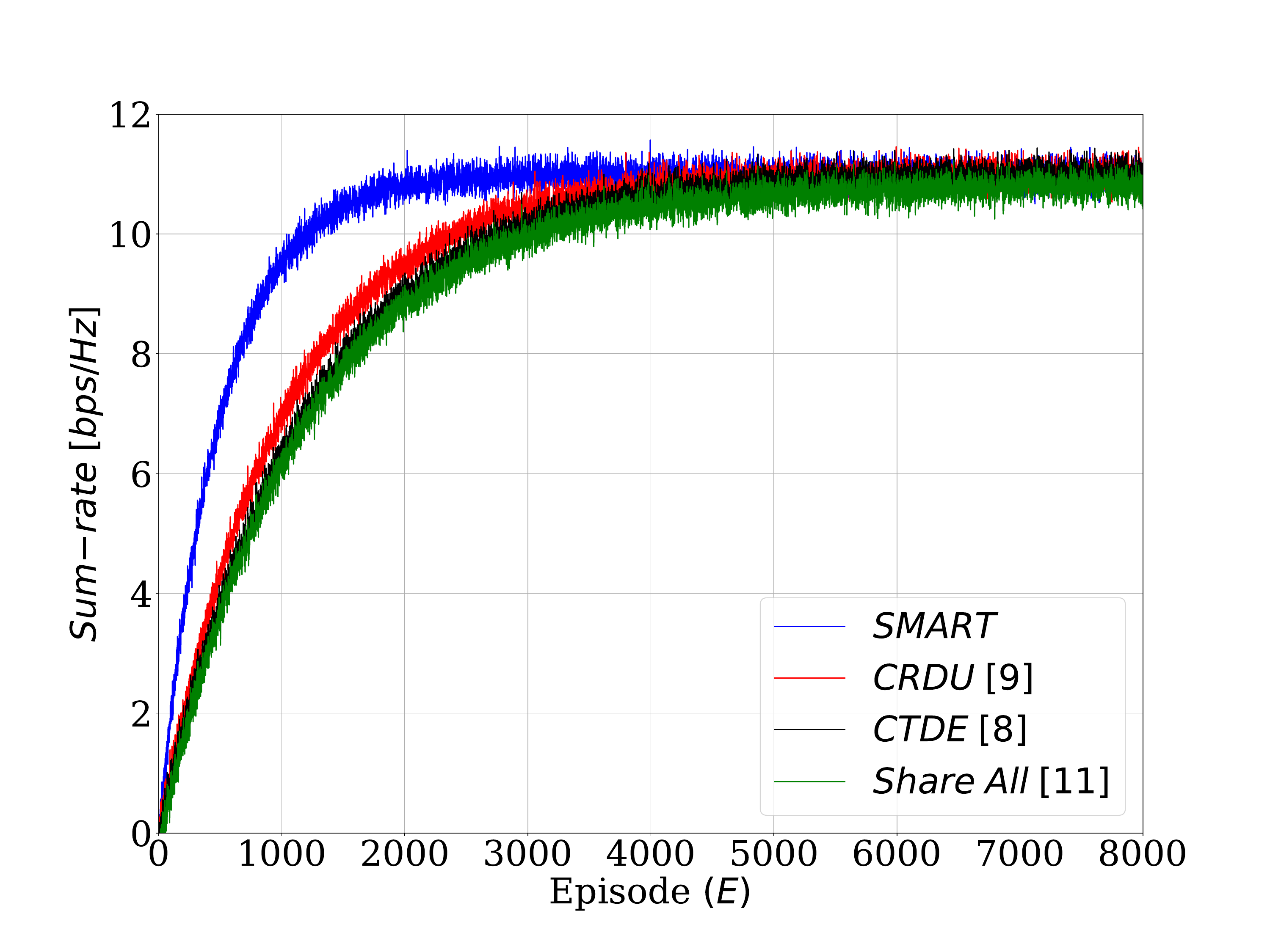}
		\caption{Convergence for different algorithms.}
		\label{fig:converge}
	\end{figure}

	\section{Simulation Results}\label{sec:trainResult}

 \subsection{Simulation Setup}\label{sec:Setup}
    We consider a multi-cell network operating in the mmwave spectrum with hexagonal geometry each with a cell radius of $112m$ and inter-site distance of $225m$. The operation frequency is 28 GHz. UEs are uniformly distributed and are moving at a speed of 2 km/h. We have $\gamma_{\min}   = -3$ dB, $L=2$, $U=3$ and $\Re=100$.  We are considering Rayleigh fading, where signal strength undergoes random variations following a Rayleigh distribution. The DQN network parameters are $\alpha = 0.995$, $\eta = 0.01$ and $B = 32$. All networks have two hidden layers with 56 neurons and ReLU activation function. 

To evaluate our algorithm, we compare it with several approaches. \textit{{CTDE}} (based on \cite{lu2021dynamic}) uses a central agent whose weight is shared among agents. The \textit{{CRDU}} (based on \cite{zhang2022multi})  framework maximizes the system sum-rate by performing local network updates while a central controller uniformly dictates rewards/penalties. \textit{{Multi-agent baselines}} include ``Share Nothing,” where agents do not share experiences, and ``Share All,” \cite{dahal2023multi} where all experiences are shared among agents. 

Spectral efficiency (measured by achievable sum-rate) is
the main performance evaluation measure. We evaluate the
average network sum-rate by
{\begin{align}\label{eq:Sum_rate__}
		{{R_{\rm sum}}}   = {\dfrac{1}{E}\sum_{e=1}^{E}\sum_{\ell = 1 }^L\sum_{u=1}^{U} \log_2(1+\gamma_{\ell,u}^{[e]})},
		\end{align}}where $E$ is the total number of episodes within which the agent
interacts with the environment, $\gamma_{\ell,u}^{[e]}$ is the effective SINR at episode $e$.   Another performance measure is overall network coverage, evaluated by the
\textit{cumulative distribution function (CCDF) }of the effective SINR of all users.

\begin{algorithm} 
  \caption{Training phase of the proposed algorithm}
  \begin{algorithmic}[1]\label{alg:alg1}
  \STATE Initialize $Q_\ell(\mathbf{s}_{\ell,t}, {\mathbf{a}_{\ell,t}}), \forall L$ with random weights $\bm\theta_{t},\forall L$
  \STATE Initialize local reply buffer  $R_\ell, \forall L$ 
  \STATE  for episode 1 to $E$ do
    \STATE \quad for $t$=1 to $T$ do
    \STATE \quad \quad for $\ell$=1 to $L$ do
    \STATE \quad \quad \quad Observe local state $\bm s_{\ell,t}$
    \STATE \quad \quad \quad 
    Compute local action based on~\eqref{eq:action}, rewards based \\ 
    \quad \quad \quad 
    on ~\eqref{eq:Sum_rate__} and observe the next local state $\bm s_{\ell,t+1}$
    \STATE \quad \quad \quad {Store transition $  (\mathbf{s}_{\ell,t},\mathbf{a}_{\ell,t},r_{\ell,t},\mathbf{s}_{\ell,t+1})$ in $R_\ell$ }
    \STATE \quad \quad  end for
    \STATE \quad \quad for $\ell$=1 to $L$ do
    \STATE \quad \quad \quad Select experiences $\bm c_{\ell,j,t}$ based on~\eqref{eq:comm}
    \STATE \quad \quad \quad for each agent $j \neq \ell$ do
    \STATE \quad \quad \quad \quad Insert $\bm c_{\ell,j,t}$ into buffer $R_j$ 
    \STATE \quad \quad \quad end for
    \STATE \quad \quad end for
    \STATE \quad \quad for $\ell$=1 to $L$ do
    \STATE \quad \quad \quad {Perform GD on~\eqref{eq:Loss} and update $\bm\theta_{t}$} 
    \STATE \quad \quad end for
    \STATE \quad \quad $s_{\ell,t} = s_{\ell,t+1}$
    \STATE \quad end for
    \STATE end for
  \end{algorithmic}
\end{algorithm}

 \subsection{Results}\label{sec:Results}

 We first compare the performance of the proposed algorithm with different algorithms as shown in Fig.~\ref{fig:cmpSNRR1}.
 In Fig.~\ref{a}, the CCDFs of effective SINR for different algorithms are compared. The proposed SMART algorithm result is close to that of the CTDE and ``Share All'' algorithms. The CTDE and ``Share All'' algorithms perform better as they take advantage of having the complete set of experiences shared among agents. The ``Share Nothing'' algorithm performs poorly because it lacks experience sharing between BSs, which is crucial for effective interference mitigation. Similarly, CRDU also shows reduced performance because its central controller imposes uniform rewards/ penalties, which can be limiting as the poor performance of a single agent impacts all agents.
With the proposed SMART algorithm, 30\% of the time UEs achieve $\rm SINR > 20$ dB.  This shows the algorithm's effectiveness in enhancing network performance and managing interference by utilizing a few relevant experiences from other BSs.

Figure~\ref{b} shows the network sum-rate for the different algorithms. It is noticeable that the performance of the proposed algorithm (SMART) is close to that of the CTDE and ``Share All'' algorithms and is better than ``Share Nothing'' and CRDU algorithms. {We compare our SMART algorithm with three non-RL-based algorithms: the global CSI scheme \cite{mismar2019deep}, which optimizes transmit power and beamforming for all UEs in each BS using full CSI; the ICIC scheme \cite{kim2020sum}, which improves the sum-rate by sharing limited SINR information among BSs to reduce inter-cell interference; and the signal-to-leakage-plus-noise ratio (SLNR) scheme \cite{li2022decentralized}, which minimizes interference to other UEs while maintaining signal quality for the target UE. SMART almost matches the global CSI scheme's performance as few pertinent
experiences from other agents can lead to effective learning. In contrast, ICIC's reliance on limited SINR sharing can be restrictive, as poor performance in one cell may negatively impact others, and the SLNR scheme, while distributed, underperforms because it does not directly optimize individual UEs' signal quality.} Lastly, in Fig.~\ref{c}, we can see that about 75\% of the time, no experiences are shared between the BSs. This outcome is particularly significant because with only 25\% of experiences shared, the performance of the proposed algorithm is reasonablly high. This shows that even a few relevant experiences from other BSs can achieve almost same performance of ``Share All'' and CTDE algorithms. 

In Fig.~\ref{fig:converge}, we illustrate the convergence of various algorithms by plotting sum-rate versus episodes. The proposed algorithm converges faster than the others, primarily due to its selective sharing of relevant experiences from other agents. In contrast, the ``Share All'' and CTDE algorithms require a complete set of experiences shared among agents.

\section{Conclusion}\label{sec:conclu}

We have introduced a selective experience sharing multi-agent algorithm that enhances interference mitigation, aiming to maximize the network sum-rate. Our approach is based on inter-cell interference power, a useful metric for quantifying the interference caused by neighboring BSs to the serving BS. 
Experiences are shared among cells based on the inter-cell interference power.  
Simulation results demonstrate improved performance compared to the multi-agent algorithms (Share Nothing and CRDU) and comparable performance to the baseline algorithms (Share All and CTDE). Our proposed scheme minimizes the per-BS experience sharing, making it dependent solely on the inter-cell interference power of users rather than sharing all experiences. The effectiveness of our proposed scheme is verified through numerical simulations.

 \typeout{}
	\bibliography{SMART_Algorithm}
	\bibliographystyle{ieeetr}

\end{document}